\documentclass[11pt,a4paper]{article}
\usepackage[hyperref]{acl2021}
\usepackage{times}
\usepackage{latexsym}
\usepackage{booktabs}
\usepackage{graphicx}
\usepackage{url}

\usepackage{microtype}

\aclfinalcopy 

\title{ERICA: An Empathetic Android Companion for Covid-19 Quarantine}

\author{Etsuko Ishii$^1$, Genta Indra Winata$^1$, Samuel Cahyawijaya$^1$, Divesh Lala$^2$, \\\textbf{Tatsuya Kawahara}$^2$, \textbf{Pascale Fung}$^{1}$ \\
  $^1$Center for Artificial Intelligence Research (CAiRE), HKUST\\
  $^2$Graduate School of Informatics, Kyoto University \\
  \texttt{\{eishii, giwinata, scahyawijaya\}@connect.ust.hk}
  }

\date{}

\begin{document}
\maketitle
\begin{abstract}
Over the past year, research in various domains, including Natural Language Processing (NLP), has been accelerated to fight against the COVID-19 pandemic, yet such research has just started on dialogue systems.
In this paper, we introduce an end-to-end dialogue system which aims to ease the isolation of people under self-quarantine.
We conduct a control simulation experiment to assess the effects of the user interface, a web-based virtual agent called Nora vs. the android ERICA via a video call.
The experimental results show that the android offers a more valuable user experience by giving the impression of being more empathetic and engaging in the conversation due to its nonverbal information, such as facial expressions and body gestures. Demo video available at \url{https://youtu.be/PLPEBXLeKJI}.
\end{abstract}

\section{Introduction}
To combat the COVID-19 pandemic, lockdowns have been imposed around the world, leading many to experience social isolation.
Many people have also undergone weeks of mandatory self-quarantine as they crossed a border or had close contact with a patient.
The resulting social loneliness can affect people's mental state, and mental support for those under isolation is suggested~\citep{choi2020depression, zhao2020social}. 
For more than half a century, dialogue systems have played the role of therapist, psychologist or counselor~\citep{vaidyam2019chatbots},
and many were designed to help people with a specific concern~\citep{rizzo2011simcoach, devault2014simsensei}.
Hence, dialogue systems have a role in helping curb the effects of social isolation arising from the pandemic.


\begin{figure*}[t]
    \centering
    \includegraphics[width=0.85\textwidth]{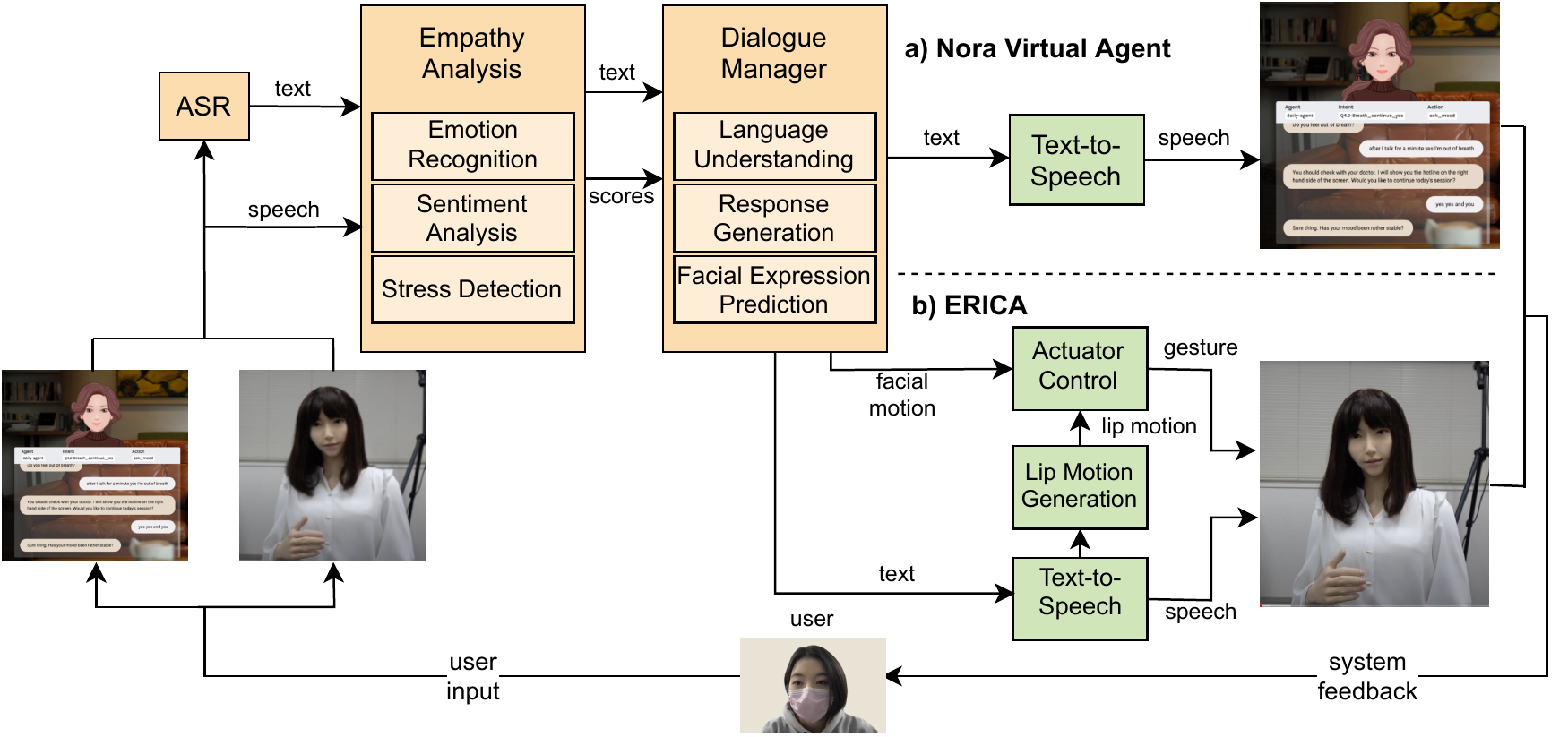}
    \caption{Architecture of a spoken dialogue system for the (a) Nora web-based virtual agent and (b) android ERICA. Note that modules coloured in orange are shared, while the others differ depending on the agent type.}
    \label{fig:system_flow}
    \vspace{-10pt}
\end{figure*}

\begin{figure}[!t]
    \centering
    \includegraphics[width=0.6\linewidth]{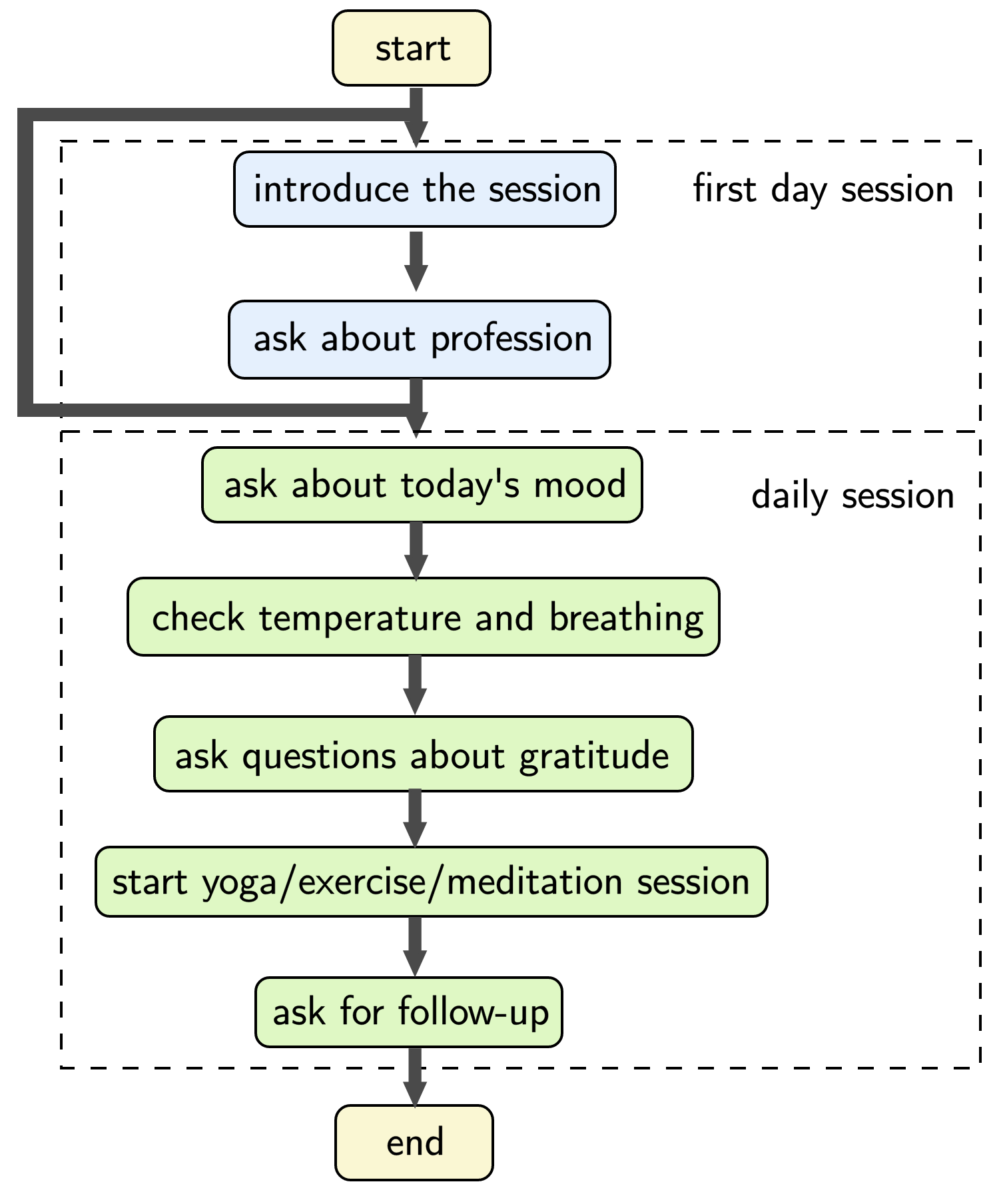}
    \vspace{-5pt}
    \caption{General dialogue flow of Nora.}
    \label{fig:dialogue-flow}
    \vspace{-15pt}
\end{figure}

To meet the emerging needs arising from the pandemic, we extend the idea of Nora, an empathetic dialogue system which mimics a conversation with a psychologist~\citep{winata2017nora, winata2021nora}, to specifically mentally support people under self-quarantine, and we install her dialogue system into the autonomous android ERICA~\cite{glas2016erica}. We utilize ERICA's nonverbal features, which are not offered by Nora, to improve the user interface (UI), because it is well-accepted that the nonverbal behavior of clinicians and therapists affects the outcome of patients~\citep{foley2010nonverbal, beck2002physician}.
During the conversation session, our system asks a set of questions to screen for stress and depression as well as health conditions such as body temperature or shortness of breath.
We conduct a comparative study of the virtual agents between the web-based Nora and android ERICA, and we design a dialogue flow particularly for quarantined users based on Nora's graphical UI. 

The experimental results show that nonverbal information actually enhances the quality of the user experience during the session by giving the user the impression he or she is being empathized with and listened to.
This suggests the importance of the design of nonverbal behavior in dialogue agents,
especially for those in the mental health care domain.

\section{Conversational Agents}
Here we describe the end-to-end system for the Nora web-based virtual agent and the android ERICA, whose architectures are depicted in Figure~\ref{fig:system_flow}. 

\subsection{Dialogue Manager}

The dialogue manager consists of three sub-modules: language understanding, response generation, and facial expression prediction. The language understanding module detects the user's intent and slot entities. The response generation module will then generate an appropriate response sentence according to the information from language understanding and empathy analysis. The system utterance generated from the response generation is then passed to the facial expression prediction module, which decides the appropriate facial expression to show. The facial expression is categorized into six distinct classes: happiness, sadness, anger, surprise, laughter, and neutral.

We design a dialogue flow that focuses on a conversation with users in quarantine.
As shown in Figure~\ref{fig:dialogue-flow}, Nora's dialogue conversation is divided into two sessions, the first day session and daily session. In the first day session, the agent will introduce the session and ask about the user's profession.
The agent will proceed in the daily sessions by asking about the user's mood and continue with a temperature and shortness of breath check. Afterward, the agent asks questions about gratitude and then recommends that the user enjoy activities such as yoga, exercise, and meditation. At the end of each activity, the agent will ask a follow-up question about how the user feels about the activity. When ending the conversation, the agent will say goodbye and remind the user to wash their hands and wear a mask.

\subsection{Empathy Analysis}
The empathy analysis module contains three modules to understand the user's mood: stress detection,
sentiment analysis, and emotion recognition from text and audio~\citep{winata2017nora}. These modules are later used in the dialogue manager to respond appropriately without discomforting the user. We compute stress, sentiment, and emotion scores on every user turn, and use them to identify whether the user has an extreme psychological condition or not. We also use the scores to track the user's mood every day and provide suggestions to the user for improving their mental well-being.

\subsection{Nora Virtual Agent}
The Nora virtual agent has a web interface, as shown in Figure~\ref{fig:system_flow}a, that accepts speech input. Users can see their input and responses in text as well as the automatic speech recognition (ASR) results of their utterances. To improve the interaction, the virtual agent provides sound effects to signal the user when the system starts and stops listening.
To make the conversation more natural, Nora uses a text-to-speech (TTS) module to generate a speech response. 

\subsection{ERICA}
\begin{figure}[t]
    \centering
    \includegraphics[width=0.80\linewidth]{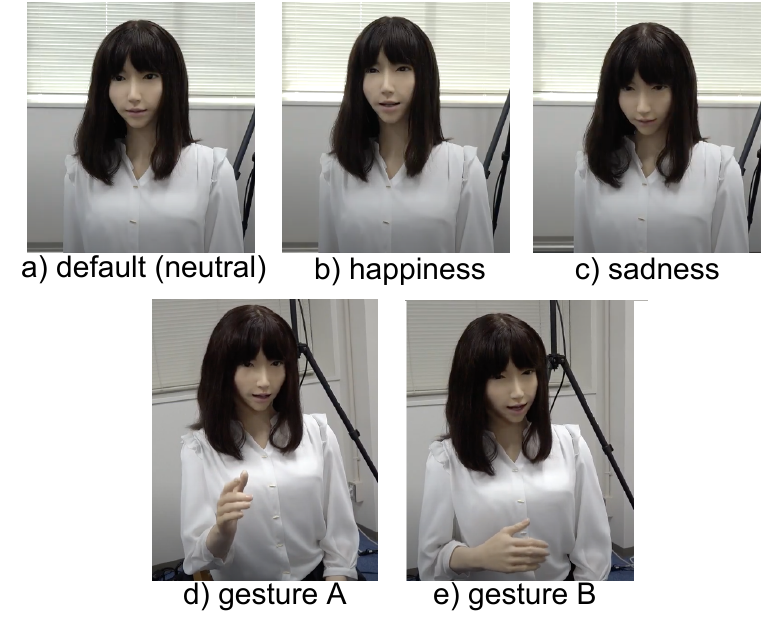}
    \vspace{-10pt}
    \caption{ERICA's facial expressions and gestures.}
    \label{fig:nonverbal_erica}
    \vspace{-10pt}
\end{figure}

ERICA is a super-realistic female humanoid developed as a conversational agent to play various roles~\citep{glas2016erica}.
She has facial expressions controlled by a facial expression predictor inside the dialogue manager.
We develop a mapping of the emotion category to ERICA's actual facial movement and execute it during her utterance, with examples shown in Figures~\ref{fig:nonverbal_erica}(a), (b), and (c).
During the user turn, ERICA adopts the default (neutral) face.
She also has a lip-motion generation module which is directly controlled by speech signals obtained by the TTS module.


We implement nonverbal behaviors which are triggered based on the turn: body gestures during the system turn, and nodding during the user turn. Body gestures are intended to show openness to users during ERICA's utterance, mainly moving her right hand, as shown in Figure~\ref{fig:nonverbal_erica}(d) and (e).
We design four versatile movements and play one of them randomly during ERICA's utterance.
During the user turn, ERICA nods to play the role of an active listener until 2.0~s of user silence is detected.

To enhance the naturalness of ERICA's behavior during the conversation, a random gazing model is also introduced.
ERICA normally does speaker tracking using Kinect~\citep{inoue-etal-2016-talking, inoue-etal-2020-attentive}, but since the participant in our case is not on-site, we model gazing behavior as a random uniform sampling of a gaze point nearby the webcam.
The gaze point will be randomly changed within a hollow cylinder from the center of the webcam with an outer radius of 0.3~m, inner radius of 0.05~m, and width of 0.2~m. The gaze change decision is taken every 1.5~s.

\section{Experiments}
We conducted a comparative evaluation to see how nonverbal information such as facial expressions and body gestures affect the user experience by asking volunteers to participate in a session with the Nora virtual agent and ERICA.

\subsection{Experimental Setup}
We conducted a simulation of counseling and recruited 19 participants who are fluent in English.
In the experiment, a participant accessed the web interface through their web browser and reached the dialogue session page as in Figure~\ref{fig:system_flow}(a) to have a session with Nora. 
Then, using a video conference tool, they talk with ERICA
just as they would a usual video call.

After finishing the two sessions,
we asked participants to evaluate the two systems by choosing which agent is preferred from four different criteria based on their experience during the conversation.
Participants were also asked to give an additional comment describing the reason for their choice on each criterion.

\subsection{Results and Analysis}

\begin{table}[!th]
    \centering
    \resizebox{0.49\textwidth}{!}{
    \begin{tabular}{lcc}
    \toprule
    \multicolumn{1}{c}{Question Item} & ERICA & Nora Virtual Agent \\ \midrule
    Q1. Overall Experience & 52.6 & 47.4 \\
    Q2. Empathy & \textbf{68.4} & 31.6 \\
    Q3. Attentiveness & \textbf{94.7} & 5.3 \\
    Q4. User Friendliness & 21.1 & \textbf{78.9} \\ \bottomrule    
    \end{tabular}
    }
    \caption{Human evaluation results in terms of the winning rate (\%) with participants of $n=19$. Bold denotes statistically significant (one-sided t-test with $p < 0.1$). }
    \label{tab:results}
\end{table}

In Table~\ref{tab:results}, we summarize the experimental results. 
Overall, ERICA is only slightly preferred (52.6\%) over the Nora virtual agent (47.4\%) due to its system drawbacks, even though it is perceived to be more attentive and empathetic. 

\textbf{Q1: Overall Experience} is comparable for several reasons: Although ERICA is regarded as more empathetic and engaging in conversations, users reported that they had a poorer experience, mainly because of the delay in ERICA's response. Moreover, some participants pointed out that the virtual agent is preferable since calling ERICA every day might be troublesome.

\textbf{Q2: Empathy} shows that ERICA is perceived as significantly more empathetic thanks to its facial expressions and gestures. 
Some participants reported that gestures reflected their emotions and thus ERICA was being empathetic, even though her gestures are independent of their emotions.

\textbf{Q3: Attentiveness} shows that ERICA is perceived to be significantly more attentive to users because of her nodding, facial expressions, and gestures that mimic human listening behaviors to some extent. Most of the participants agreed that the feedback from ERICA during the user turn, namely, nodding, reduced their anxiety about not being understood. 

\textbf{Q4: User Friendliness} measures technical or psychological difficulties. 
The majority of the participants reported that ASR accuracy and response time are the drawbacks of ERICA, while some preferred ERICA as she is more human-like and easier to talk to.
To enhance the user friendliness, further investigation should be done to handle additional environmental noise in the video call.

\section{Related Work}
One of the major challenges in dialogue systems is how to incorporate empathy, and several papers have explored approaches for end-to-end chatbots~\citep{lin2020caire, ma2020survey}.
Empathetic dialogue systems are attracting more interest in the field of psychiatry as well~\citep{vaidyam2019chatbots}, especially those equipped with nonverbal features~\citep{devault2014simsensei, rizzo2011simcoach}.
In addition,~\citet{inoue-etal-2016-talking, inoue-etal-2020-attentive} utilized ERICA's nonverbal features to make her more empathetic in more generic situations. 

\section{Conclusion}
In this paper, we described the implementation of the Nora dialogue system and its application in the android ERICA. A comparison of ERICA against Nora shows that the facial expressions and body gestures of ERICA give a better impression of attentiveness and empathy, even though ERICA has technical drawbacks such as delayed response and worse ASR quality than Nora. These results suggest that nonverbal communication is crucial for machine-to-human conversation as for human-to-human conversation, and special care is needed to design the nonverbal behaviors of empathetic dialogue systems.

\bibliographystyle{acl_natbib}
\bibliography{acl2021}



\end{document}